\documentclass[runningheads]{llncs}
\usepackage{graphicx}
\usepackage{tikz}
\usepackage{comment}
\usepackage{amsmath,amssymb} % define this before the line numbering.
\usepackage{color}
\usepackage{orcidlink}

\usepackage{enumitem}
\usepackage{color}
\usepackage{xcolor}
\usepackage{comment}
\usepackage{graphicx}
\usepackage{multirow}
\usepackage{placeins}

\newcommand{\smallsec}[1]{\noindent {\bf #1.}}
\usepackage[capitalize]{cleveref}
\crefname{section}{Sec.}{Secs.}
\Crefname{section}{Section}{Sections}
\Crefname{table}{Table}{Tables}
\crefname{table}{Tab.}{Tabs.}
\begin{document}
% \linenumbers
\pagestyle{headings}
\mainmatter
\def\ECCVSubNumber{1511}  % Insert your submission number here

\title{HIVE: Evaluating the Human \\Interpretability of Visual Explanations} % Replace with your title

\begin{comment}
\titlerunning{ECCV-22 submission ID \ECCVSubNumber} 
\authorrunning{ECCV-22 submission ID \ECCVSubNumber} 
\author{Anonymous ECCV submission}
\institute{Paper ID \ECCVSubNumber}
\end{comment}
%******************

%\begin{comment}
\titlerunning{HIVE: Evaluating the Human Interpretability of Visual Explanations}
\author{Sunnie S. Y. Kim\orcidlink{0000-0002-8901-7233}\index{Kim, Sunnie S. Y.}
\and
Nicole Meister\orcidlink{0000-0002-7154-6882}
\and
Vikram V. Ramaswamy\orcidlink{0000-0002-0552-5338}\index{Ramaswamy, Vikram V.}
\and
\\
Ruth Fong\orcidlink{0000-0001-8831-6402}
\and
Olga Russakovsky\orcidlink{0000-0001-5272-3241}
}
\authorrunning{Kim, Meister, Ramaswamy, Fong, Russakovsky}
\institute{Princeton University, Princeton NJ 08544, USA
\\
\email{\{sunniesuhyoung, nmeister, vr23, ruthfong, olgarus\}@princeton.edu}
}
%\end{comment}
%******************

\maketitle

\begin{abstract}
As AI technology is increasingly applied to high-impact, high-risk domains, there have been a number of new methods aimed at making AI models more human interpretable. Despite the recent growth of interpretability work, there is a lack of systematic evaluation of proposed techniques. In this work, we introduce HIVE (Human Interpretability of Visual Explanations), a novel human evaluation framework that assesses the utility of explanations to human users in AI-assisted decision making scenarios, and enables falsifiable hypothesis testing, cross-method comparison, and human-centered evaluation of visual interpretability methods. To the best of our knowledge, this is the first work of its kind. Using HIVE, we conduct IRB-approved human studies with nearly 1000 participants and evaluate four methods that represent the diversity of computer vision interpretability works: GradCAM, BagNet, ProtoPNet, and ProtoTree. Our results suggest that explanations engender human trust, even for incorrect predictions, yet are not distinct enough for users to distinguish between correct and incorrect predictions. We open-source HIVE to enable future studies and encourage more human-centered approaches to interpretability research.
HIVE can be found at \url{https://princetonvisualai.github.io/HIVE}.
\end{abstract}

\section{Introduction}
\label{sec:intro}

With the growing adoption of AI in high-impact, high-risk domains, there have been a surge of efforts aimed at making AI models more interpretable. 
Motivations for interpretability include allowing human users to trace through a model’s reasoning process (accountability, transparency), verify that the model is basing its predictions on the right reasons (fairness, ethics), and assess their level of confidence in the model (trustworthiness).
The \textit{interpretability} research field tackles these questions and is comprised of diverse works, including those that provide explanations of the behavior and inner workings of complex AI models~\cite{bau2017netdissect,bau2019seeing,fong2019extremal,fong2018net2vec,petsiuk2018rise,selvaraju2017gradcam,simonyan2013deep,zeiler2014visualizing,Zhou2016CAM}, those that design inherently interpretable models~\cite{brendel2019bagnet,Boehle2021CVPR,Boehle2022CVPR,chen2019protopnet,donnelly2022deformable,dubey2022scalable,koh2020concept,nauta2021prototree,radenovic2022neural}, and those that seek to understand what is easy and difficult for these models~\cite{agarwal2020estimating,Wang_2018_ECCV,alert2014cvpr} to make their behavior more interpretable.

Despite much methods development, there is a relative lack of standardized evaluation methods for proposed techniques.
Existing evaluation methods for computer vision interpretability methods are focused on feature attribution heatmaps that highlight ``important'' image regions for a model's prediction.
Since we lack ground-truth knowledge about which regions are \textit{actually} responsible for the prediction, different evaluation metrics use different proxy tasks for verifying these important regions (e.g., measuring the impact of deleting regions or the overlap between ground-truth objects and highlighted regions)~\cite{fong2017meaningful,hooker2019roar,petsiuk2018rise,Poppi_2021_CVPR,yang2019benchmarking,zhang2016EB}.
However, these automatic evaluation metrics are disconnected from downstream use cases of explanations; they don't capture how useful end-users find heatmaps in their decision making.
Further, these metrics don't apply to other forms of explanations, such as prototype-based explanations produced by some of the recent interpretable-by-design models~\cite{chen2019protopnet,donnelly2022deformable,nauta2021prototree}.

\begin{figure}[t!]
\centering
\includegraphics[width=\linewidth]{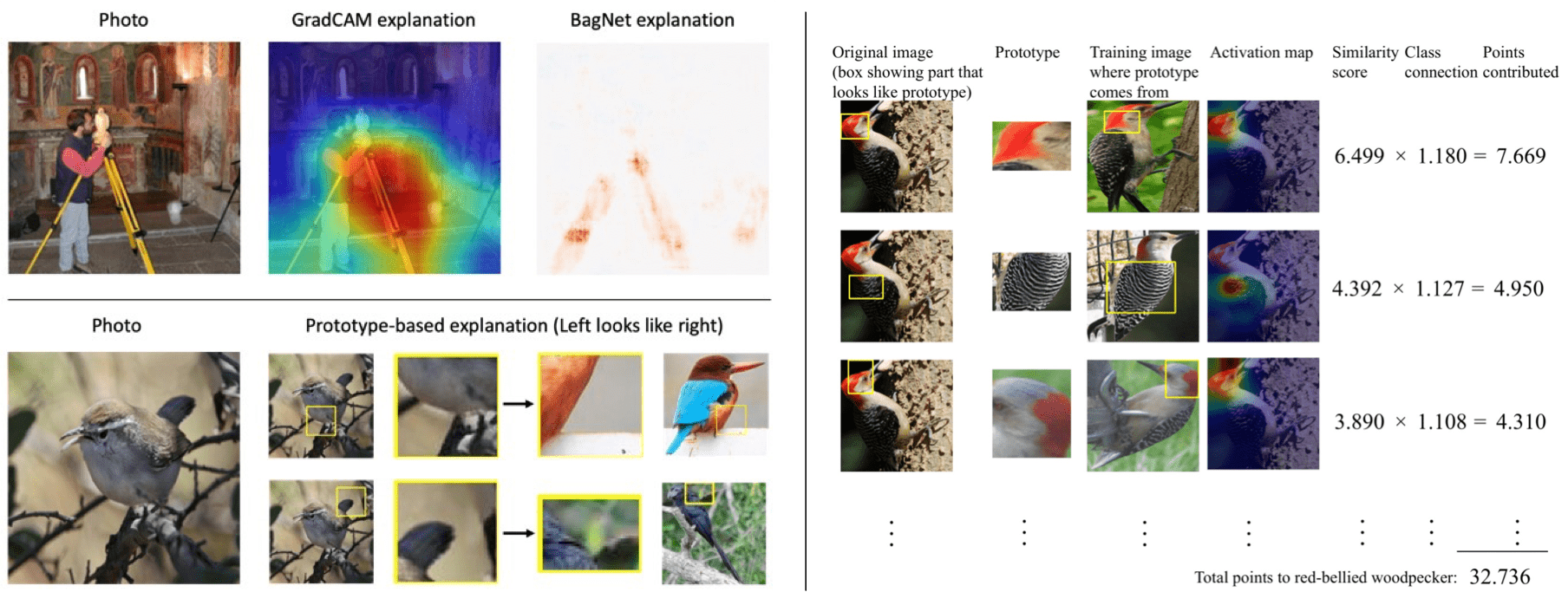}
\caption{\textbf{Different forms of explanation.}
(Top left) Heatmap explanations (GradCAM~\cite{selvaraju2017gradcam}, BagNet~\cite{brendel2019bagnet}) highlight decision-relevant image regions.
(Bottom left) Prototype-based explanations (ProtoPNet~\cite{chen2019protopnet}, ProtoTree~\cite{nauta2021prototree}) match image regions to prototypical parts learned during training.
This schematic is much simpler than actual explanations.
(Right) Actual ProtoPNet explanation example from the original paper.
\textbf{While existing evaluation methods typically apply to only one explanation form, HIVE evaluates and compares diverse interpretability methods.}
}
\label{fig:explanations}
\end{figure}

In part due to these challenges, the interpretability of a proposed method is often argued through a few exemplar explanations that highlight how a method is more interpretable than a baseline model.
However, recent works suggest that some methods are not as interpretable as originally imagined and may engender over-trust in automated systems~\cite{adebayo2018neurips,dzindolet2003trust,herlocker2000recommend,hoffmann2021looks,lipton2017mythos,margeloiu2021concept,nguyen2021neurips,shen2020hcomp}.
They caution against an over-reliance on intuition-based justifications and raise awareness for the need of falsifiable hypotheses~\cite{leavitt2020falsifiable} and proper evaluation in interpretability research.

\smallsec{Our contributions}
As more diverse interpretability methods are being proposed, it is more important than ever to have a standardized and rigorous evaluation framework that allows for falsifiable hypothesis testing, cross-method comparison, and human-centered evaluation.
To this end, we develop HIVE (Human Interpretability of Visual Explanations).
HIVE evaluates diverse visual interpretability methods by evaluating all methods on a common task.
We carefully design the tasks to reduce the effect of confirmation bias and human prior knowledge in interpretability evaluation, and assess the utility of explanations in AI-assisted decision making scenarios.
HIVE also examines how well interpretable-by-design models' reasoning process aligns with that of humans, and how human users tradeoff interpretability and accuracy.

To demonstrate the extensibility and applicability of HIVE, we conduct IRB-approved human studies with nearly 1000 participants and evaluate four existing methods that represent different streams of interpretability work (e.g., post-hoc explanations, interpretable-by-design models, heatmaps, and prototype-based explanations): GradCAM~\cite{selvaraju2017gradcam}, BagNet~\cite{brendel2019bagnet}, ProtoPNet~\cite{chen2019protopnet}, ProtoTree~\cite{nauta2021prototree}.
To the best of our knowledge, we are the first to compare interpretability methods with different explanation forms (see~\cref{fig:explanations}) and the first to conduct human studies of the evaluated interpretable-by-design models~\cite{brendel2019bagnet,chen2019protopnet,nauta2021prototree}.

We obtain a number of insights through our studies:
\begin{itemize}[topsep=0pt]
\item When provided explanations, participants tend to believe that the model predictions are correct, revealing an issue of \emph{confirmation bias}. For example, our participants found 60\% of the explanations for \emph{incorrect} model predictions convincing. Prior work has made similar observations for non-visual interpretability methods~\cite{poursabzi2021manipulating}; we substantiate them for visual explanations and demonstrate a need for rigorous evaluation of proposed methods.
\item When given multiple model predictions and explanations, participants struggle to distinguish between correct and incorrect predictions based on the explanations (e.g., achieving only 40\% accuracy on a multiple-choice task with four options). 
This result suggests that interpretability methods need to be improved to be reliably useful for AI-assisted decision making.
\item There exists a gap between the similarity judgments of humans and prototype-based models~\cite{chen2019protopnet,nauta2021prototree} which can hurt the quality of their interpretability.
\item Participants prefer to use a model with explanations over a baseline model without explanations. To switch their preference, they require the baseline model to have $+6.2\%$ to $+10.9\%$ higher accuracy.
\end{itemize}

As interpretability is fundamentally a human-centric concept, it needs to be evaluated in a human-centric way. We hope our work helps pave the way towards human evaluation becoming commonplace, by presenting and analyzing a human study design, demonstrating its effectiveness and informativeness for interpretability evaluation, and open-sourcing the code to enable future work.

\section{Related work}
\label{sec:relatedwork}

\smallsec{Interpretability landscape in computer vision}
Interpretability research can be described along several axes: first, whether a method is post-hoc or interpretable-by-design; second, whether it is global or local; and third, the form of an explanation (see \cite{arrieta2019explainable,brundage2020trustworthy,chen2021survey,fong2020thesis,gilpin2018explaining,Gunning_Aha_2019,rudin2021survey,samek2019book} for surveys).
\emph{Post-hoc explanations} focus on explaining predictions made by already-trained models, whereas \emph{interpretable-by-design (IBD)} models are intentionally designed to possess a more explicitly interpretable decision-making process~\cite{brendel2019bagnet,Boehle2021CVPR,Boehle2022CVPR,chen2019protopnet,donnelly2022deformable,dubey2022scalable,koh2020concept,nauta2021prototree,radenovic2022neural}. 
Furthermore, explanations can either be \emph{local explanations} of a single input-output example or \emph{global explanations} of a network (or its component parts).
Local, post-hoc methods include heatmap~\cite{fong2019extremal,petsiuk2018rise,selvaraju2017gradcam,shitole2021sag,simonyan2013deep,zeiler2014visualizing,Zhou2016CAM}, counterfactual explanation~\cite{goyal2019counterfactual,vandenhende2022counterfactual,wang2020scout}, approximation~\cite{ribeiro2016lime}, and sample importance~\cite{koh2017influence,yeh2018representer} methods.
In contrast, global, post-hoc methods aim to understand global properties of CNNs, often by treating them as an object of scientific study~\cite{bau2017netdissect,bau2019seeing,fong2018net2vec,kim2020gestalt} or by generating class-level explanations~\cite{ramaswamy2022elude,zhou2018ibd}.
Because we focus on evaluating the utility of explanations in AI-assisted decision making, we do not evaluate global, post-hoc methods. 
\emph{IBD} models can provide local and/or global explanations, depending on the model type.
Lastly, explanations can take a variety of forms: two more popular ones we study are \emph{heatmaps} highlighting important image regions and \emph{prototypes} (i.e., image patches) from the training set that form interpretable decisions.
In our work, we investigate four popular methods that span these types of interpretability work: GradCAM~\cite{selvaraju2017gradcam} (post-hoc, heatmap), BagNet~\cite{brendel2019bagnet} (IBD, heatmap), ProtoPNet~\cite{chen2019protopnet} (IBD, prototypes), and ProtoTree~\cite{nauta2021prototree} (IBD, prototypes).
See~\cref{fig:explanations} for examples of their explanations.

\smallsec{Evaluating heatmaps}
Heatmap methods are arguably the most-studied class of interpretability work.
Several automatic evaluation metrics have been proposed~\cite{bach2015pixel,fong2017meaningful,hooker2019roar,petsiuk2018rise,Poppi_2021_CVPR,yang2019benchmarking,zhang2016EB}, however, there is a lack of consensus on how to evaluate these methods.
Further, the authors of~\cite{adebayo2018neurips,adebayo2020neurips} and BAM~\cite{yang2019benchmarking} highlight how several methods fail basic ``sanity checks'' and call for more comprehensive metrics.
Complementing these works, we use HIVE to study how useful heatmaps are to human users in AI-assisted decision making scenarios and demonstrate insights that cannot be gained from automatic evaluation metrics.

\smallsec{Evaluating interpretable-by-design models}
In contrast, there has been relatively little work on assessing interpretable-by-design models.
Quantitative evaluations of these methods typically focus on demonstrating their competitive performance with a baseline CNN, while the quality of their interpretability is often demonstrated through qualitative examples.
Recently, a few works revisited several methods' interpretability claims.
Hoffmann et al.~\cite{hoffmann2021looks} highlight that prototype similarity of ProtoPNet~\cite{chen2019protopnet} does not correspond to semantic similarity and that this disconnect can be exploited.
Margeloiu et al.~\cite{margeloiu2021concept} analyze concept bottleneck models~\cite{koh2020concept} and demonstrate that learned concepts fail to correspond to real-world, semantic concepts.
In this work, we conduct the first human study of three popular interpretable-by-design models~\cite{brendel2019bagnet,chen2019protopnet,nauta2021prototree} and quantify prior work's~\cite{hoffmann2021looks,nauta2021prototree} anecdotal observation on the misalignment between prototype-based models~\cite{chen2019protopnet,nauta2021prototree} and humans' similarity judgment.

\smallsec{Evaluating interpretability with human studies}
Outside the computer vision field, human studies are commonly conducted for models trained on tabular datasets~\cite{lage2019human,lage2018human,lakkaraju2016,poursabzi2021manipulating,zhang2020fat}; however, these do not scale to the complexity of modern vision models.
Early human studies for visual explanations have been limited in scope: They typically ask participants which explanation they find more reasonable or which model they find more trustworthy based on explanations~\cite{jeyakumar2020neurips,selvaraju2017gradcam}.
Recently, more diverse human studies have been conducted~\cite{biessmann2019psychophysics,borowski2021exemplary,fel2021evaluation,nguyen2021neurips,shen2020hcomp,shitole2021sag,zimmermann2021causal}.

Closest to our work are~\cite{fel2021evaluation,nguyen2021neurips,shen2020hcomp}.
Shen and Huang~\cite{shen2020hcomp} ask users to select incorrectly predicted labels with or without showing explanations; Nguyen et al.~\cite{nguyen2021neurips} ask users to decide whether model predictions are correct based on explanations; Fel et al.~\cite{fel2021evaluation} ask users to predict model outputs in a concurrent work.
Regarding \cite{nguyen2021neurips,shen2020hcomp}, our \textit{distinction} task also investigates how useful explanations are in distinguishing correct and incorrect predictions. However, different from these works, we ask users to select the correct prediction out of multiple predictions to reduce the effect of confirmation bias and don't show class labels to prevent users from relying their prior knowledge. 
Regarding \cite{fel2021evaluation}, we also ask users to predict model outputs, but mainly as a supplement to our \textit{distinction} task. Further, we ask users to identify the model output out of multiple predictions based on the explanations, whereas \cite{fel2021evaluation} first trains users to be a meta-predictor of the model by showing example model predictions and explanations, and then at test time asks users to predict the model output for a given image without showing any explanation.
Most importantly, different from \cite{fel2021evaluation,nguyen2021neurips,shen2020hcomp}, we evaluate interpretability methods beyond heatmaps and conduct cross-method comparison.
Our work is similar in spirit to work by Zhou et al.~\cite{zhou2019hype} on evaluating generative models with human perception.
For general guidance on running human studies in computer vision, refer to work by Bylinskii et al.~\cite{bylinskii2022userstudies}.

\section{HIVE design principles}
\label{sec:framework}

In this work, we focus on AI-assisted decision making scenarios, in particular those that involve an image classification model. 
For a given input image, a user is shown a model’s prediction along with an associated explanation, and is asked to make a decision about whether the model's prediction is correct or more generally about whether to use the model.
In such a scenario, explanations are provided with several goals in mind: help the user identify if the model is making an error, arrive at a more accurate prediction, understand the model’s reasoning process, decide how much to trust the model, etc.

To study whether and to what extent different visual interpretability methods are useful for AI-assisted decision making, we develop a novel human evaluation framework named HIVE (Human Interpretability of Visual Explanations).
In particular, we design HIVE to allow for \textit{falsifiable hypothesis testing} regarding the usefulness of explanations for identifying model errors, \textit{cross-method comparison} between different explanation approaches, and \textit{human-centered evaluation} for understanding the practical effectiveness of interpretability.

\begin{figure*}[t!]
\centering
\includegraphics[width=\linewidth]{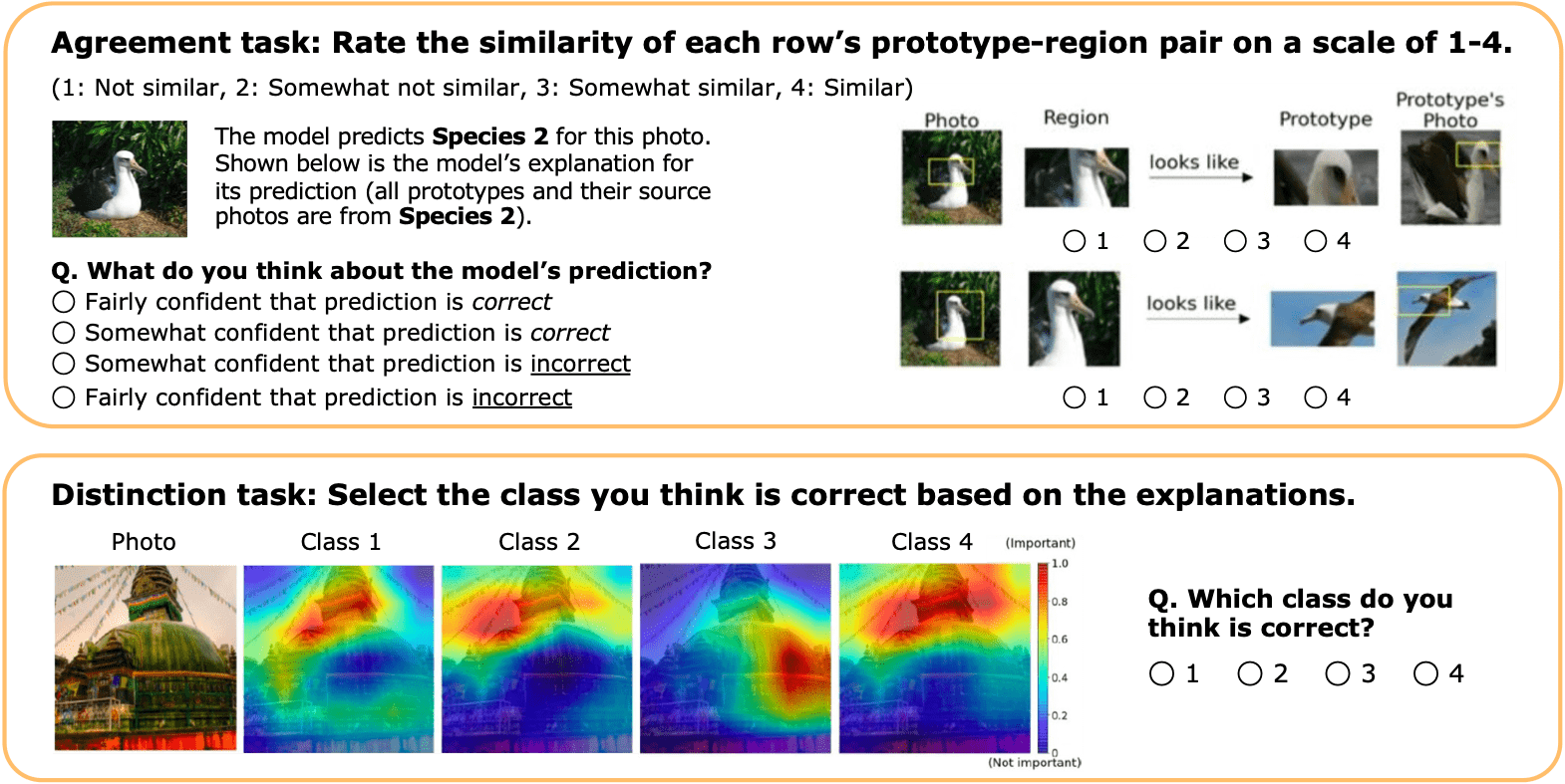}
\caption{\textbf{Study user interfaces (UIs).}
We show simplified UIs for evaluating ProtoPNet~\cite{chen2019protopnet} on the \emph{agreement} task (top) and GradCAM~\cite{selvaraju2017gradcam} on the \emph{distinction} task (bottom).
Full UI snapshots are in supp. mat. See \cref{sec:framework} for description of the tasks.
}
\label{fig:ui}
\end{figure*}

\subsection{Falsifiable hypothesis testing}
\label{sec:hypothesistesting}

We join a growing body of work that cautions against intuition-based justification and subjective self-reported ratings in interpretability evaluation~\cite{adebayo2018neurips,leavitt2020falsifiable,Kunkel2019trust,Schaffer2019trust} and calls for objective assessment with behavior indicators~\cite{Lai2019trust,poursabzi2021manipulating,Yin2019trust,zhang2020fat}.
To this end, we design two evaluation tasks, the \textit{agreement} and \textit{distinction} tasks, that enable \textit{falsifiable hypothesis testing} about the evaluated interpretability method.

In the \textit{agreement} task, we present participants with one prediction-explanation pair at a time and ask how confident they are in the model's prediction based on the explanation.
We evaluate methods on this task in part because it is closer to existing interpretability evaluation schemes that consider a model's top-1 prediction and its explanation~\cite{selvaraju2017gradcam}, and also because it allows us to quantify the degree to which participants believe in model predictions based on explanations.

The \textit{agreement} task measures the amount of \textit{confirmation bias} that arises for a given interpretability method. However, it doesn't measure the utility of explanations in distinguishing correct and incorrect predictions, a crucial functionality of explanations in AI-assisted decision making.
Hence, we design and use the \textit{distinction} task as our main evaluation task.
Here we simultaneously show four predictions and their associated explanations for a given input image and ask users to identify the correct prediction based on the provided explanations.
The \textit{distinction} task also mitigates the effect of confirmation bias in interpretability evaluation, as participants now have to reason about multiple explanations at once.
See~\cref{fig:ui} for the evaluation task UIs.

One concern with this setup is ensuring that participants use the provided explanations rather than their knowledge to complete the task.
We take two measures to remove the effect of \textit{human prior knowledge} in our evaluations.
First, we evaluate all interpretability methods in the context of fine-grained bird species classification~\cite{WahCUB_200_2011}, which is a challenging task for non-bird experts. 
Second, as a more general measure, we omit the semantic class labels of the predictions.
This measure is particularly important when evaluating interpretability methods in easier contexts, e.g., coarse-grained object classification with ImageNet~\cite{ILSVRC15}, because the task becomes too easy otherwise (i.e., participants can select the correct prediction based on the class labels instead of using the explanations).
Note that ground-truth class labels are also omitted to simulate a realistic decision making scenario where users do not have access to the ground truth.

\subsection{Cross-method comparison}
\label{sec:crossmethod}

Existing evaluation methods typically apply to only one explanation form (e.g., heatmaps are compared against each other).
In contrast, HIVE enables \textit{cross-method comparison} between different explanation forms by focusing on downstream uses of explanations and evaluating all methods on a common task.

However, there remains a number of practical roadblocks. First, different methods may have been developed for different scenarios (e.g., fine-grained vs. coarse-grained classification), requiring us to carefully analyze the effect of the particular setting during evaluation.
Second, different methods may be more or less digestible to the users. While this is an inherent part of what we are trying to evaluate, we also want to ensure that the evaluation task is doable by study participants with limited machine learning background, given most human studies in the field are run through Amazon Mechanical Turk.
Hence, we actualize a specific evaluation setup for each interpretability method by creating an individual evaluation UI that respects the method's characteristics (e.g., its explanation form, dataset used for model training).
We briefly describe the four methods we evaluate in this work (see \cref{fig:explanations} for example explanations) and their evaluation setups.
When making any adaptations, we tried to present each method in as favorable of a way as possible.
More details are in supp. mat.

\smallsec{GradCAM~\cite{selvaraju2017gradcam}}
GradCAM is a post-hoc method that produces a heatmap that highlights important regions in an input image that contribute to a model's prediction.
We evaluate GradCAM on ImageNet~\cite{ILSVRC15}, which it was originally developed for, as well as on CUB~\cite{WahCUB_200_2011}, for which we train a standard CNN model to use as the underlying model for generating GradCAM heatmaps.

\smallsec{BagNet~\cite{brendel2019bagnet}}
In contrast, BagNet is an interpretable-by-design model that collects evidence for a class from small regions of an image.
For each class, BagNet creates a heatmap where higher values (i.e., darker red in our visualizations) imply stronger evidence for the class.
BagNet then sums the values in each heatmap and predicts the class with the highest sum.
We evaluate BagNet on ImageNet, for which it was originally designed, as well as on CUB, for which we train a new BagNet model using the authors' code.

\smallsec{ProtoPNet~\cite{chen2019protopnet}} 
The next two methods reason with \emph{prototypes}, which are small image patches from the training set that these models deem as representative for a certain class. 
At test time, ProtoPNet compares a given image to the set of prototypes it learned during training and finds regions in the image that are the most similar to each prototype.
It computes a similarity score between each prototype-region pair, then predicts the class with the highest weighted sum of the similarity scores.
The ProtoPNet model for CUB learns 10 prototypes for each of the 200 bird species (2,000 total) and produces one of the most complex explanations.
Its explanation for a single prediction consists of 10 prototypes and their source images, heatmaps that convey the similarity between matched image regions and prototypes, continuous and unnormalized similarity scores, and weights multiplied to the scores (see~\cref{fig:explanations} right).
In our evaluation, we abstract away most technical details based on our pilot studies, and focus on showing the most crucial component of ProtoPNet's reasoning process: the prototype-image region matches.
We also ask participants to rate the similarity of each match (see~\cref{fig:ui} top) to assess how well the model's similarity judgment aligns with that of humans.
See supp. mat. for the task and explanation modification details.

\smallsec{ProtoTree~\cite{nauta2021prototree}} 
Finally, the ProtoTree model learns a tree structure along with the prototypes.
Each node in the tree contains a prototype from a training image. 
At each node, the model compares a given test image to the node's prototype and produces a similarity score.
If the score is above some threshold, the model judges that the prototype is present in the image and absent if not.
The model then proceeds to the next node and repeats this process until it reaches a leaf node, which corresponds to a class. 
The ProtoTree model for CUB trained by the authors has 511 decision nodes and up to 10 decision steps, and our pilot studies revealed that is too overwhelming for participants.
Thus in our evaluation, we significantly simplify the decision process. 
Participants are shown the model's decisions until the penultimate decision node, and then are asked to make decisions for only the final two nodes of the tree by judging whether the prototype in each node is present or absent in the image.
This leads the participants to select one of the four ($2^2$) classes as the final prediction.
One additional challenge is that participants may not be familiar with decision trees and thus may have trouble following the explanation. 
To help understanding, we introduce a simple decision tree model with two levels, walk through an example, and present two warm up exercises so that participants can get familiar with decision trees before encountering ProtoTree. See supp. mat. for more information.

\subsection{Human-centered evaluation}
\label{humancentered}

HIVE complements existing algorithmic evaluation methods by bringing humans back into the picture and taking a \textit{human-centered} approach to interpretability evaluation.
The design of HIVE, particularly the inclusion/exclusion of class labels in Sec.~\ref{sec:hypothesistesting} and careful actualization of the evaluation setup in Sec.~\ref{sec:crossmethod}, is focused on making this evaluation tractable for the participants and as fair as possible with respect to different interpretability methods. 
We also went through multiple iterations of UI design to present visual explanations in digestible bits so as to not overwhelm participants with their complexity.
Despite the challenges, there is a very important payoff from human studies. We are able to evaluate different interpretability methods through participants' 1) ability to \emph{distinguish} between correct and incorrect predictions based on the provided explanations, simulating a more realistic AI-assisted decision-making setting, and 2) level of \emph{alignment} with the model's intermediate reasoning process in the case of prototype-based, interpretable-by-design models. 
We also gain a number of valuable insights that can only be obtained through human studies.

\subsection{Generalizability \& Scalability}
\label{generalizability}

In closing we discuss two common concerns about human studies: generalizability and scalability.
We have shown HIVE's \textit{generalizability} by using it to evaluate a variety of methods (post-hoc explanations, interpretable-by-design models, heatmaps, prototype-based explanations) in two different settings (coarse-grained object recognition with ImageNet, fine-grained bird recognition with CUB).
Further, a recent work by Ramaswamy et al.~\cite{ramaswamy2022roadblocks} uses HIVE to set up new human studies, for evaluating example-based explanations and finding the ideal complexity of concept-based explanations, demonstrating that HIVE can be easily generalized to new methods and tasks.
Regarding \textit{scalability}, human study costs are not exorbitant contrary to popular belief and can be budgeted for like we budget for compute. For example, our GradCAM distinction study cost \$70 with 50 participants compensated at \$12/hr. The real obstacles are typically the time, effort, and expertise required for study design and UI development; with HIVE open-sourced, these costs are substantially mitigated.

\section{HIVE study design}
\label{sec:studydesign}

In this section, we describe our IRB-approved study design.
See supp. mat. and \url{https://princetonvisualai.github.io/HIVE} for UI snapshots and code.

\smallsec{Introduction}
For each participant, we first introduce the study and receive their informed consent.
We also request optional demographic data regarding gender identity, race and ethnicity, and ask about the participant's experience with machine learning; however, no personally identifiable information was collected.
Next we explain the evaluated interpretability method in simple terms by avoiding technical jargon (i.e., replacing terms like ``image'' and ``training set'' to ``photo'' and ``previously-seen photos'').
We then show a preview of the evaluation task and provide example explanations for one correct and one incorrect prediction made by the model to give the participant appropriate references. 
The participant can access the method description at any time during the task.

\smallsec{Objective evaluation tasks}
Next we evaluate the interpretability method on a behavioral task (\textit{distinction} or \textit{agreement}) introduced in~\cref{sec:hypothesistesting} and~\cref{fig:ui}.
Detailed task descriptions are available in supp. mat.

\smallsec{Subjective evaluation questions} 
While the core of HIVE is in the objective evaluation tasks, we also ask subjective evaluation questions to make the most out of the human studies.
Specifically, we ask the participant to self-rate their level of understanding of the evaluated method before and after completing the task, to investigate if the participant's self-rated level of understanding undergoes any changes during the task. After the task completion, we disclose the participant's performance on the task and ask the question one last time.

\smallsec{Interpretability-accuracy tradeoff questions}
While interpretability methods offer useful insights into a model's decision, some explanations come at the cost of lower model accuracy.
Hence in the final part of the study, we investigate the \textit{interpretability-accuracy tradeoff} participants are willing to make when comparing an interpretable method against a baseline model that doesn't come with any explanation.
In high-risk scenarios a user may prefer to maximize model performance over interpretability.
However, another user may prefer to prioritize interpretability in such settings so that there would be mechanisms for examining the model's predictions. 
To gain insight into the tradeoff users are willing to make, we present three scenarios: low-risk (e.g., bird species recognition for scientific or educational purposes), medium-risk (e.g., object recognition for automatic grocery checkout), and high-risk (e.g., scene understanding for autonomous driving). 
For each scenario, we then ask the participant to input the minimum accuracy of the baseline model that would convince them to use it over the model with explanations and also describe the reason for their choices.

\section{Experiments}
\label{sec:experiments}

\subsection{Experimental details}

\smallsec{Datasets \& Models}
We evaluate all interpretability methods on classification tasks and use images from the CUB~\cite{WahCUB_200_2011} test set and the ImageNet~\cite{ILSVRC15} validation set to generate model predictions and explanations.
On CUB, we evaluate all four methods: GradCAM~\cite{selvaraju2017gradcam}, BagNet~\cite{brendel2019bagnet}, ProtoPNet~\cite{chen2019protopnet}, ProtoTree~\cite{nauta2021prototree}.
On ImageNet, we evaluate GradCAM and BagNet.
See supp. mat. for details.

\smallsec{Human studies}
For each study, i.e., an evaluation of one interpretability method on one task (\textit{distinction} or \textit{agreement}), we recruited 50 participants through Amazon Mechanical Turk (AMT).
In total, we conducted 19 studies with 950 participants; see supp. mat. for the full list.
The self-reported machine learning experience of the participants was $2.5 \pm 1.0$, between ``2: have heard about...'' and ``3: know the basics...''
The mean study duration was 6.9 minutes for GradCAM, 6.6 for BagNet, 13.6 for ProtoPNet, and 10.4 for ProtoTree.
Participants were compensated based on the state-level minimum wage of \$12/hr.

\smallsec{Statistical analysis}
For each study, we report the mean task accuracy and standard deviation of the participants' performance which captures the variability between individual participants' performance.
We also compare the study result to random chance and compute the $p$-value from a 1-sample $t$-test.\footnote{
We compare our results to chance performance instead of a baseline without explanations because we omit semantic class labels to remove the effect of human prior knowledge (see~\cref{sec:hypothesistesting}); so such a baseline would contain no relevant information.}
When comparing results between two groups, we compute the $p$-value from a 2-sample $t$-test. Results are deemed statistically significant under $p<0.05$ conditions.

\begin{table}[t!]
\centering
\caption{
\textbf{Agreement task results.}
For each study, we show mean accuracy, standard deviation of the participants' performance, and mean confidence rating in parentheses. \emph{Italics} denotes methods with accuracy not statistically significantly different from 50\% random chance ($p>0.05$); \textbf{bold} denotes the highest performing method in each group.
\textbf{In all studies, participants leaned towards believing that model predictions are correct when provided explanations, regardless of if they are actually correct.} For example, for GradCAM on CUB, participants thought 72.4\% of correct predictions were correct and $100 - 32.8 = 67.2\%$ of incorrect predictions were correct. These results reveal an issue of \textit{confirmation bias}. See~\cref{sec:confirmationbias} for a discussion.
}
\resizebox{\linewidth}{!}{
\begin{tabular}{|c|c|c|c|c|}
\hline
CUB & GradCAM \cite{selvaraju2017gradcam} & BagNet \cite{brendel2019bagnet} & ProtoPNet \cite{chen2019protopnet} & ProtoTree \cite{nauta2021prototree} \\
\hline
Correct & 72.4\% $\pm$ 21.5 (2.9) & \textbf{75.6\% $\pm$ 23.4 (3.0)} & 73.2\% $\pm$ 24.9 (3.0) & 66.0\% $\pm$ 33.8 (2.8) \\
Incorrect & 32.8\% $\pm$ 24.3 (2.8) & \textit{42.4\% $\pm$ 28.7 (2.7)} & \textbf{\textit{46.4\% $\pm$ 35.9 (2.4)}} & 37.2\% $\pm$ 34.4 (2.7) \\
\hline
ImageNet & GradCAM \cite{selvaraju2017gradcam} & BagNet \cite{brendel2019bagnet} & - & - \\
\hline
Correct & \textbf{70.8\% $\pm$ 26.6 (2.9)} & 66.0\% $\pm$ 27.2 (2.8) & - & - \\
Incorrect & \textbf{\textit{44.8\% $\pm$ 31.6 (2.7)}} & 35.6\% $\pm$ 26.9 (2.7) & - & - \\
\hline
\end{tabular}
}
\label{tab:agreement}
\end{table}

\subsection{The issue of confirmation bias}
\label{sec:confirmationbias}
Let us first examine how the four methods perform on the \textit{agreement} task, where we present participants with one prediction-explanation pair at a time and ask how confident they are in the model's prediction.
Results are summarized in \cref{tab:agreement}.
On CUB, participants found 72.4\% of correct predictions convincing for GradCAM, 75.6\% for BagNet, 73.2\% for ProtoPNet, and 66.0\% ProtoTree.
However, they also thought 67.2\% of incorrect predictions were correct for GradCAM, 57.6\% for BagNet, 53.6\% for ProtoPNet, and 62.8\% for ProtoTree.
Similarly on ImageNet, participants found 70.8\% of correct predictions convincing for GradCAM and 66.0\% for BagNet, yet also believed in 55.2\% and 64.4\% of incorrect predictions, respectively.
These results reveal an issue of \textit{confirmation bias}: When given explanations, participants tend to believe model predictions are correct, even if they are wrong.
Still, the confidence ratings are overall higher for correct predictions than incorrect predictions, suggesting there is some difference between their explanations.
More results and discussion are in supp. mat.

\subsection{Objective assessment of interpretability}
\label{sec:distinction}

Next we discuss findings from our main evaluation task, the \textit{distinction} task, where we ask participants to select the correct prediction out of four options based on the provided explanations.
Results are summarized in \cref{tab:distinction}.

\smallsec{Participants perform better on correctly predicted samples}
On correctly predicted samples from CUB, the mean task accuracies are 71.2\% on GradCAM, 45.6\% on BagNet, 54.5\% on ProtoPNet and 33.8\% on ProtoTree, all above the 25\% chance baseline.
That is, participants can identify which of the four explanations correspond to the ground-truth class correctly predicted by the model.
On incorrect predictions, however, the accuracies drop from 71.2\% to 26.4\% for GradCAM and from 45.6\% to 32.0\% for BagNet, and we observe a similar trend in the ImageNet studies. 
These results suggest that explanations for correct predictions may be more coherent and convincing than those for incorrect predictions.
Even so, all accuracies are far from 100\%, indicating that the evaluated methods are not yet reliably useful for AI-assisted decision making.

\begin{table}[t!]
\centering
\caption{
\textbf{Distinction and output prediction task results.}
For each study, we report the mean accuracy and standard deviation of the participants' performance.
\emph{Italics} denotes methods that do not statistically significantly outperform 25\% random chance ($p>0.05$); \textbf{bold} denotes the highest performing method in each group.
In the top half, we show the results of all four methods on CUB. In the bottom half, we show GradCAM and BagNet results on ImageNet, without vs. with ground-truth class labels.
\textbf{Overall, participants struggle to identify the correct prediction or the model output based on explanations.}
See~\cref{sec:distinction} for a discussion.
}
\resizebox{\linewidth}{!}{
\begin{tabular}{|c|c|c|c|c|c|}
\hline
\multicolumn{2}{|c|}{CUB} & GradCAM~\cite{selvaraju2017gradcam} & BagNet~\cite{brendel2019bagnet} & ProtoPNet \cite{chen2019protopnet} & ProtoTree \cite{nauta2021prototree} \\
\hline
\multirow{2}{*}{Distinction} & Correct & \textbf{71.2\% $\pm$ 33.3} & 45.6\% $\pm$ 28.0 & 54.5\% $\pm$ 30.3 & 33.8\% $\pm$ 15.9 \\
 & Incorrect & \emph{26.4\% $\pm$ 19.8} & \textbf{32.0\% $\pm$ 20.8} & - & - \\
\hline
\multirow{2}{*}{Output prediction} & Correct & \textbf{69.2\% $\pm$ 32.3} & 50.4\% $\pm$ 32.8 & - & - \\
 & Incorrect & \textbf{53.6\% $\pm$ 27.0} & \emph{30.0\% $\pm$ 24.1} & - & - \\
\hline
\multicolumn{2}{|c|}{ImageNet} & GradCAM~\cite{selvaraju2017gradcam} & with labels & BagNet~\cite{brendel2019bagnet} & with labels \\
\hline
\multirow{2}{*}{Distinction} & Correct & \textbf{51.2\% $\pm$ 24.7} & 49.2\% $\pm$ 30.8 & 38.4\% $\pm$ 28.0 & 34.8\% $\pm$ 27.7 \\
 & Incorrect & \textbf{\emph{30.0\% $\pm$ 22.4}} & \emph{27.2\% $\pm$ 20.3} & \emph{26.0\% $\pm$ 18.4} & \emph{27.2\% $\pm$ 18.7} \\
\hline
\multirow{2}{*}{Output prediction} & Correct& \textbf{48.0\% $\pm$ 28.3} & \textbf{48.0\% $\pm$ 35.6} & 46.8\% $\pm$ 29.0 & 42.8\% $\pm$ 27.4 \\
 & Incorrect & \textbf{35.6\% $\pm$ 24.1} & 33.2\% $\pm$ 25.2 & 34.0\% $\pm$ 24.1 & 32.8\% $\pm$ 25.5 \\
\hline
\end{tabular}
}
\label{tab:distinction}
\end{table}

\smallsec{Participants struggle to identify the model's prediction}
For GradCAM and BagNet, we ask participants to select the class they think the model predicts (\textit{output prediction}) in addition to the class they think is correct (\textit{distinction}).
For BagNet, this is a straightforward task where participants just need to identify the most activated (most red, least blue) heatmap among the four options, as BagNet by design predicts the class with the most activated heatmap.
However, accuracy is not very high, only marginally above the \textit{distinction} task accuracy.
This result suggests that BagNet heatmaps for the top-4 (or top-3 plus ground-truth) classes look similar to the human eye, and may not be suitable for assisting humans with tasks that involve distinguishing one class from another.
For GradCAM, participants also struggle on this task but to a lesser degree.

\smallsec{Showing ground-truth labels hurts performance}
For GradCAM and BagNet, we also investigate the effect of showing ground-truth class labels for the presented images.
We have not been showing them to simulate a realistic decision making scenario where users don’t have access to the ground truth.
However, since the task may be ambiguous for datasets like ImageNet whose images may contain several objects, we run a second version of the ImageNet studies showing ground-truth class labels on the same set of images and compare results.
Somewhat surprisingly, we find that accuracy decreases, albeit by a small amount, with class labels.
One possible explanation is that class labels implicitly bias participants to value heatmaps with better localization properties, which could be a suboptimal signal for the \textit{distinction} and \textit{output prediction} tasks.

\smallsec{Automatic evaluation metrics correlate poorly with human study results}
We also analyze GradCAM results using three automatic metrics that evaluate the localization quality of post-hoc attribution maps: pointing game~\cite{zhang2016EB}, energy-based pointing game~\cite{wang2020score}, and intersection-over-union~\cite{Zhou2016CAM}.
In the \textit{agreement} studies, we find near-zero correlation between participants' confidence in the model prediction and localization quality of heatmaps. In the \textit{distinction} studies, we also do not see meaningful relationships between the participants' choices and these automatic metrics.
These observations are consistent with the findings of \cite{nguyen2021neurips,fel2021evaluation}, i.e., automatic metrics poorly correlate with human performance in post-hoc attribution heatmap evaluation.
See supp. mat. for details.

\subsection{A closer examination of prototype-based models}

We are the first to conduct human studies of ProtoPNet and ProtoTree which produce some of the most complex visual explanations.
As such, we take a closer look at their results to better understand how human users perceive them.

\smallsec{A gap exists between similarity ratings of ProtoPNet \& ProtoTree and those of humans}
We quantify prior work's~\cite{hoffmann2021looks,nauta2021prototree} anecdotal observation that there exists a gap between model and human similarity judgment.
For ProtoTree, the Pearson correlation coefficient between the participants' similarity ratings and the model similarity scores is 0.06, suggesting little to no relationship.
For ProtoPNet, whose similarity scores are not normalized across images, we compute the Spearman's rank correlation coefficient ($\rho=-0.25, p=0.49$ for \textit{distinction} and $\rho=-0.52$, $p=0.12$ for \textit{agreement}). There is no significant negative correlation between the two, indicating a gap in similarity judgment that may hurt the models' interpretability.
See supp. mat. for more discussion.

\smallsec{Participants perform relatively poorly on ProtoTree, but they understand how a decision tree works}
Since the previously described ProtoTree \textit{agreement} study does not take into account the model's inherent tree structure, we run another version of the study where, instead of asking participants to rate each prototype's similarity, we ask them to select the first step they disagree with in the model's explanation. The result of this study ($52.8\% \pm 19.9\%$) is similar to that of the original study ($53.6\% \pm 15.2\%$); in both cases, we cannot conclude that participants outperform 50\% random chance ($p = 0.33$, $p = 0.10$). To ensure participants understand how decision trees work, we provided a simple decision tree example and subsequent questions asking participants if the decision tree example makes a correct or incorrect prediction. Participants achieved $86.5\%$ performance on this task, implying that the low task accuracy for ProtoTree is not due to a lack of comprehension of decision trees. See supp. mat. for details.

\subsection{Subjective evaluation of interpretability}

To complement the objective evaluation tasks, we asked participants to self-rate their level of method understanding three times.
The average ratings are $3.7 \pm 0.9$ after the method explanation, $3.8 \pm 0.9$ after the task, and $3.5 \pm 1.0$ after seeing their task performance, which all lie between the fair (3) and good (4) ratings.
Interestingly, the rating tends to \emph{decrease} after participants see their task performance ($p<0.05$). 
Several participants indicated that their performance was lower than what they expected, whereas no one suggested the opposite, suggesting that participants might have been disappointed in their task performance, which in turn led them to lower their self-rated level of method understanding.

\subsection{Interpretability-accuracy tradeoff}

In the final part of our studies, we asked participants for the minimum accuracy of a baseline model they would require to use it over the evaluated interpretable model with explanations for its predictions.
Across all studies, participants require the baseline model to have a higher accuracy than the model that comes with explanations, and by a greater margin for higher-risk settings.
On average, participants require the baseline model to have $+6.2\%$ higher accuracy for low-risk, $+8.2\%$ for medium-risk, and $+10.9\%$ for high-risk settings.
See supp. mat. for the full results and the participants' reasons for their choices.

\section{Conclusion}
\label{sec:conclusion}

In short, we introduce and open-source HIVE, a novel human evaluation framework for evaluating diverse visual interpretability methods, and use it to evaluate four existing methods: GradCAM, BagNet, ProtoPNet, and ProtoTree.

There are a few limitations of our work:
First, we use a relatively small sample size of 50 participants for each study due to our desire to evaluate four methods, some under multiple conditions.
Second, while HIVE takes a step towards use case driven evaluation, our evaluation setup is still far from real-world uses of interpretability methods.
An ideal evaluation would be contextually situated and conducted with domain experts and/or end-users of a real-world application (e.g., how would bird experts choose to use one method over another when given multiple interpretability methods for a bird species recognition model).

Nonetheless, we believe our work will facilitate more user studies and encourage human-centered interpretability research~\cite{Ehsan2020HCXAI,Ehsan2021HCXAI,Ehsan2022HCXAI,Liao2021HCXAI}, as our human evaluation reveals several key insights about the field.
In particular, we find that participants generally believe model predictions are correct when given explanations for them.
Humans are naturally susceptible to confirmation bias; thus, interpretable explanations will likely engender trust from humans, even if they are incorrect.
Our findings underscore the need for evaluation methods that fairly and rigorously assess the usefulness and effect of explanations.
We hope our work helps shift the field's objective from focusing on method development to also prioritizing the development of high-quality evaluation methods.

\smallsec{Acknowledgments}
This material is based upon work partially supported by the National Science Foundation (NSF) under Grant No. 1763642. Any opinions, findings, and conclusions or recommendations expressed in this material are those of the author(s) and do not necessarily reflect the views of the NSF.
We also acknowledge support from the Princeton SEAS Howard B. Wentz, Jr. Junior Faculty Award (OR), Princeton SEAS Project X Fund (RF, OR), Open Philanthropy (RF, OR), and Princeton SEAS and ECE Senior Thesis Funding (NM).
We thank the authors of~\cite{brendel2019bagnet,chen2019protopnet,He2015resnet,hoffmann2021looks,nauta2021prototree,selvaraju2017gradcam} for open-sourcing their code and/or trained models.
We also thank the AMT workers who participated in our studies, anonymous reviewers who provided thoughtful feedback, and Princeton Visual AI Lab members (especially Dora Zhao, Kaiyu Yang, and Angelina Wang) who tested our user interface and provided helpful suggestions.

\clearpage
\bibliographystyle{splncs04}
\bibliography{references}

\end{document}